\documentclass[10pt,letterpaper]{article}
\usepackage[top=0.85in,left=1.25in,footskip=0.75in,marginparwidth=2in]{geometry}

\usepackage[utf8]{inputenc}
\usepackage{caption}
\usepackage{subcaption}
\usepackage{cite}
\usepackage{amsmath}
\usepackage{nameref,hyperref}

\usepackage{microtype}
\DisableLigatures[f]{encoding = *, family = * }

\usepackage{setspace} 
\usepackage[right]{lineno}

\usepackage{changepage}

\usepackage[aboveskip=1pt,labelfont=bf,labelsep=period,singlelinecheck=off]{caption}

\makeatletter
\renewcommand{\@biblabel}[1]{\quad#1.}
\makeatother

\usepackage{lastpage,fancyhdr,graphicx}
\usepackage{epstopdf}
\fancyheadoffset[L]{2.25in}
\fancyfootoffset[L]{2.25in}
\hypersetup{hidelinks}
\usepackage{color}

\definecolor{Gray}{gray}{.25}

\usepackage{graphicx}

\usepackage{sidecap}

\usepackage{wrapfig}
\usepackage[pscoord]{eso-pic}
\usepackage[fulladjust]{marginnote}
\usepackage{graphicx}
\usepackage{epstopdf}
\reversemarginpar

\begin{document}
\vspace*{0.35in}

\begin{flushleft}
{\Large
\textbf\newline{The Strange Attractor Model of Bipedal Locomotion and its Consequences on Motor Control}
}
\newline
\\
Carlo Tiseo\textsuperscript{1,2,3,4*},
Ming Jeat Foo \textsuperscript{2,3},
Kalyana C Veluvolu \textsuperscript{5},
Arturo Forner-Cordero \textsuperscript{6},
Wei Tech Ang\textsuperscript{2,3}
\\
\bigskip
\bf{1} School of Engineering and Informatics, University of Sussex, Sussex House, Falmer
Brighton, BN1 9RH, UK;\\
\bf{2} Rehabilitation Research Institute of Singapore, 50 Nanyang Avenue N3-01a-01, Singapore 639798, Singapore;
\\
\bf{3} School of Mechanical \& Aerospace Engineering, Nanyang Technological University, 50 Nanyang Avenue N3-01a-01, Singapore 639798, Singapore;
\\
\bf{4} School of Informatics, University of Edinburgh, 1.17 Bayes Centre 47 Potterrow, EH8 9BT Edinburgh, UK;
\\
\bf{5} School of Electronics Engineering,Kyungpook National University, Daegu  702701, South Korea;
\\
\bf{6} Biomechatronics Laboratory. Escola Politecnica da Universidade de Sao Paulo, 05508-030, Sao Paulo, Brasil;
\\
\bigskip
* CT : c.tiseo@sussex.ac.uk 

\end{flushleft}
\section*{Abstract}
\noindent Despite decades of research, the dynamic mechanisms governing human locomotion remain incompletely understood. Current motor control models only partially capture these complex dynamics, a theoretical gap that likely limits the clinical efficacy of modern lower-limb rehabilitation therapies. Recently, it has been proposed that human locomotion is planned in task-space by explicitly exploiting the gravitational pull acting on the Center of Mass (CoM). Building upon this concept, this paper presents a novel deterministic task-space planner for straight, constant-speed locomotion. Our model formulates the CoM trajectory in the transverse plane as a harmonic oscillator moving at a constant forward velocity, while demonstrating that the vertical CoM trajectory is actively regulated through localized ankle strategies. We derive closed-form equations that successfully generate human-like trajectories for both the CoM and the swing foot. The model's predictive accuracy is empirically validated against the kinematic behavior of 12 healthy subjects walking at three self-selected speeds. Furthermore, the proposed planner is mathematically compatible with an extended formulation of the Passive Motion Paradigm ($\lambda_0$-PMP), establishing a comprehensive theoretical foundation for decentralized, hierarchical architectures of semi-autonomous controllers in both human motor control and assistive robotics.

\subsection*{\textit{Keywords:}} \noindent Bipedal Locomotion, Motor Control, Human Locomotion
\newpage
\section*{\textit{Introduction}}
Human locomotion has been extensively studied for decades due to its clinical and biomimetic relevance. While Inman and Eberhart’s classic "six determinants" of gait (1953) \cite{inman1953major} guided clinical evaluations toward kinematic metrics—such as pelvic tilt, knee flexion, and lateral pelvic displacement—these determinants lacked a dynamical systems perspective. Concurrently, roboticists have sought to decode biological locomotion to replicate its robustness and efficiency. However, the exact mechanisms underlying animal locomotor strategies remain heavily debated \cite{duysens2019controller}, particularly given the complex interplay between the mechanical and computational performance of the human body \cite{duysens2018walking,Torricelli2016,Hogan2012, Hogan2013, Winter1995, Ivanenko2006,Kuo2007}. Existing computational models only partially capture these intricate locomotion dynamics \cite{Torricelli2016, Kuo2007}.\\
This limited dynamical understanding restricts our knowledge of gait control. One prominent framework is the motor primitives theory, which posits that human actions emerge from the superposition of basic movements \cite{Ivanenko2006, Tommasino2017, Hogan2012, Lacquaniti2012, Zelik2014}. Observed across various upper and lower limb strategies, this theory justifies how the central nervous system (CNS) manages biomechanical redundancy and neural transmission delays. Extending this concept, the Passive Motion Paradigm (PMP) frames motor control as the minimization of energetic costs \cite{Tommasino2017, Flash2016, Hogan2012}. PMP optimizes an energetic function that balances task intention with the mechanical impedance of the human body. However, PMP assumes body impedance dominates task dynamics, limiting its applicability in tasks with significant environmental dynamics, such as ground contact during gait.\\
To address environmental interactions, dynamic motion primitives were introduced to characterize the task attractor at the human-environment interface \cite{Hogan2012, Hogan2013}. This approach aligns with a hierarchical, semi-autonomous controller model: lower-level modules autonomously execute sub-tasks (e.g., navigating via-points) assigned and supervised by a higher-level controller \cite{Ahn2012, Hogan2013, Tiseo2018}. This decentralized architecture helps the CNS mitigate neural delays and coordinate full-body movement \cite{Ahn2012, Tiseo2018, Hogan2012}.\\
Alternatively, Optimal Feedback Control (OFT) models motor control through the minimization of a cost function associated with system dynamics \cite{Shadmehr2008, Tommasino2017, Hogan2012}. Despite its theoretical appeal, OFT faces two major drawbacks: the inability to accurately measure muscular cost functions in humans, and an exponentially high computational burden that scales with task complexity and kinematic redundancy \cite{Shadmehr2008, Tommasino2017, Hogan2012, Ajemian2010}.\\
From an experimental perspective, gait exhibits periodic stride-to-stride fluctuations that possess long-term, fractal-like correlations \cite{hausdorff2005gait, delignieres2009fractal}. This observation has driven the analysis of gait as a limit cycle attractor, a method frequently used to assess empirical multi-stride data and responses to perturbations \cite{forner2006describing}. The presence of strange attractors with fractal dynamics in human walking has been experimentally validated \cite{delignieres2009fractal}.\\
Recent works propose formulating locomotion dynamics via the potential energy of the bipedal system in task space, thereby removing the switching dynamics typically associated with swing-stance transitions \cite{Tiseo2016,tiseo2018bipedal, Tiseo2018,Tiseo2018bioinspired,Tiseo2018a}. Tiseo \textit{et al.} derived closed-form equations for the Center of Mass (CoM) and swing foot trajectories from the bipedal potential energy surface, which is characterized as a saddle \cite{tiseo2018bipedal}. Applying the gradient operator to this manifold reveals the energy flow, identifying the system's autonomous trajectory. By defining a reference frame centered on the saddle point and aligned with the surface's principal directions, energy conservation principles dictate the optimal trajectory along the gravitational manifold \cite{Tiseo2018bioinspired}. This saddle frame acts as an observer for the CoM trajectory, modeled as a harmonic oscillation between two potential energy maxima, before a coordinate transformation expresses the trajectories in task space. While this model accurately tracks the swing foot and the CoM in the transversal plane, it exhibits tracking errors in the vertical CoM trajectory \cite{tiseo2018bipedal}.\\
To address this limitation, this work presents a refined model that incorporates an accurate representation of heel-strike and toe-off phases to eliminate vertical CoM tracking errors. The simulated trajectories of the proposed model are validated against experimental gait data from healthy subjects, demonstrating improved predictive accuracy.

\section*{Method}
This section details the experimental data collection protocol, the extraction of biomechanical and gait-dependent parameters, and the validation of the proposed model.
\subsection*{Experimental gait data}
\begin{figure}[ht]
	\centering
	\includegraphics[width=\linewidth]{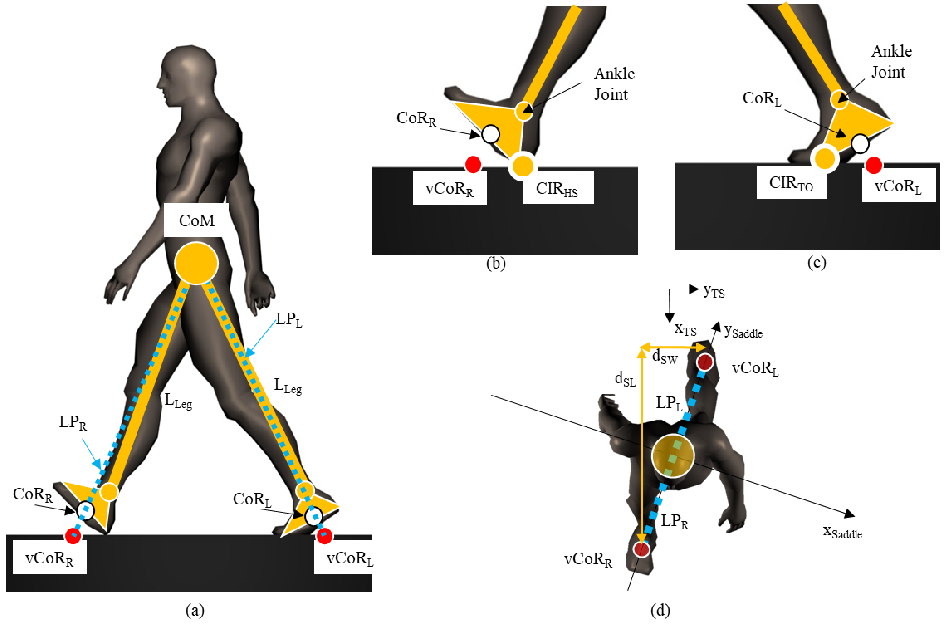}
	\caption{(a) The mechanical bipedal structure used to formulate the model consists of two identical legs connected to the CoM via spherical joints. Each leg comprises two rigid bodies (a leg segment, $L_{Leg}$, and a foot) connected via a revolute joint at the ankle. The feet are modelled as rigid triangles, with the first vertex at the ankle joint, the second coincident with the midpoint between the metatarsal joints, and the third at the heel. The Centres of Rotation ($CoR_R$ and $CoR_L$) are defined as the midpoints of the plantar segments connecting the metatarsals to the heel. The virtual CoRs ($vCoR_R$ and $vCoR_L$) are the ground projections of the CoRs used for trajectory planning, and they coincide with the physical CoRs when the feet are in full contact with the ground. The lengths of the two pendula ($LP_R$ and $LP_L$) are defined as the Cartesian distances between the CoM and the respective vCoRs. (b) The heel-strike phase is modelled as a rotation about the foot's instantaneous centre of rotation ($CIR_{HS}$). (c) The toe-off phase is modelled as a rotation about the $CIR_{TO}$. (d) The task-space planner utilizes two reference systems. The Task-Space (TS) frame is aligned with the standard anatomical planes and is used to describe the walking trajectories. The Saddle-Space (Saddle) frame is used to describe posture-dependent quantities, which are subsequently projected onto the TS frame.}
	\label{fig1}
\end{figure}

To validate the model, gait data from 12 healthy subjects (7 males and 5 females) were recorded. The participants had a mean ($\pm$ standard deviation) age of $28.3 \pm 3.42$ years, body mass of $71.14 \pm 12.6$ kg, and height of $171 \pm 10.7$ cm. Subjects were instructed to walk along a 6-metre straight path for ten trials at three self-selected speeds: slow, normal, and fast. Kinematic data were recorded using an 8-camera Vicon Motion Capture System (Vicon Motion Systems Ltd., UK), with ten reflective markers placed on the body according to \cite{Mandery2015}. Due to technical issues, five trajectory recordings at the normal speed for one subject were excluded from the analysis. The center of mass (CoM) was approximated as the centroid of four pelvic markers (left and right anterior and posterior superior iliac spines). Foot trajectories, denoted as the Centers of Rotation ($CoR_{R,L}$), were calculated by computing the barycenter of the triangle formed by the heel and the first and fifth metatarsals, projected onto the foot sole. Foot inclination relative to the ground was defined as the angle of the segment connecting the heel marker to the midpoint between the first and fifth metatarsal markers  (see \autoref{fig1}).\\
\subsection*{Derivation of the Model Parameters}
To accurately simulate locomotion strategies, the proposed model requires the identification of specific gait-dependent variables and biomechanical parameters.

The primary gait-dependent parameters—step length ($d_{SL}$), step width ($d_{SW}$), and maximum heel-strike angle ($\theta_{MaxHS}$)—were formulated as functions of walking velocity ($v_W$). These relationships were derived by fitting the mean experimental data of the 12 subjects using a first-order polynomial regression in MATLAB (MathWorks Inc., USA), yielding the following expressions:
\begin{equation}
\label{eq1}
\left\{ \begin{array}{lll}
d_{SW} = -0.009149~ v_W + 0.1072 & \text{m} & R^2 = 0.04 \\
d_{SL} = 2(0.09399~ v_W + 0.1624) & \text{m} & R^2 = 0.89 \\
\theta_{MaxHS} = -1.162~ v_W + 9.198 & \text{deg} & R^2 = 0.02
\end{array}\right.
\end{equation}
The low $R^2$ values for the heel-strike angle and step width suggest that walking velocity alone does not sufficiently capture the variance in these parameters, implying the influence of additional biomechanical or environmental factors during locomotion. Conversely, the high $R^2$ value for step length confirms that walking speed is the primary determinant of this parameter. These observations align with existing literature \cite{tiseo2018bipedal,Hak2013}, which notes that step width is highly sensitive to lateral perturbations rather than speed alone \cite{Hak2013}.\\
Furthermore, the model utilizes the following subject-specific biomechanical parameters:
\begin{itemize}
\item $h_{body} = 1.71$ m: Model body height, corresponding to the dataset mean.
\item $L_{P0} = \sqrt{0.765^2 + 0.0955^2} \times h_{body} = 0.5943 \times h_{body}$ m: Maximum pendulum length \cite{Jo2007,Virmavirta2014}.
\item $d = 0.0921$ m: Average distance of the CoRs from both the metatarsus and the heel.
\item $d_{xAH} = 0.0253 \times h_{body}$ m: Anteroposterior distance from the heel to the ankle joint \cite{Jo2007}.
\item $d_{zA} = 0.039 \times h_{body}$ m: Vertical distance from the CoR to the ankle joint \cite{Jo2007}.
\item $d_{xA} = d - d_{xAH}$ m: Anteroposterior distance from the CoR to the ankle joint.
\item $d_{xAMT} = 2d - d_{xAH}$ m: Anteroposterior distance from the metatarsus to the ankle joint.
\end{itemize}Based on these relationships, the necessary parameters to characterize the bipedal mechanism depicted in \autoref{fig1} are defined as follows:
\begin{equation}
\label{eq2}
\left\{ \begin{array}{ll}
L_{leg} = \sqrt{L_{P0}^2 - (d_{SW}/2)^2 - d_{xA}^2} - d_{zA} & \text{m} \\
D_{AH} = \sqrt{d_{zA}^2 + d_{xAH}^2} & \text{m} \\
\theta_{DAH} = \tan^{-1}(d_{zA}/d_{xAH}) & \text{deg} \\
D_{AMT} = \sqrt{d_{zA}^2 + d_{xAMT}^2} & \text{m} \\
\theta_{DMT} = \tan^{-1}(d_{zA}/d_{xAMT}) & \text{deg}
\end{array}\right.
\end{equation}
\subsection*{CoM Trajectory in the Transverse Plane}Based on the model presented by Tiseo \textit{et al.} \cite{tiseo2018bipedal}, the CoM trajectory is modeled as a harmonic oscillator moving forward at a constant speed ($v_W$) in the transverse plane, while oscillating at a frequency of $\omega_{Step}/2 = (2t_{Step})^{-1}$. The oscillation amplitude is determined by aligning the harmonic oscillator asymptote with the ordinate of the Saddle Space ($y_{Saddle}$ in \autoref{fig1}(d)) during the step-to-step transition. The Saddle Space is defined by a posture-dependent frame, as proposed by Tiseo \textit{et al.} \cite{Tiseo2016,Tiseo2018,Tiseo2018b}, to model the system's potential energy. Consequently, the amplitude $A_y = d_{SW}/(2\pi\omega_{Step}d_{SL})$ is a function of both step width ($d_{SW}$) and step length ($d_{SL}$), yielding the following kinematic expressions:
\begin{equation}
\label{eq3}
\left\{ \begin{array}{l}
x_{CoM}(t) = v_W t \\
y_{CoM}(t) = A_y \cos(\pi \omega_{Step} t)
\end{array}\right.
\end{equation}
The vertical CoM trajectory is governed by mechanical constraints—specifically, the requirement that at least one foot remains in contact with the ground. This trajectory depends on the leg posture and is further detailed at the conclusion of the ankle strategy model.
\subsection*{vCoR Anteroposterior Trajectories}
The swing foot trajectory is modeled by synchronizing its movement with the bipedal potential energy identified by the Saddle Space \cite{Tiseo2016}. This implies that if the CoM were maintained along $y_{Saddle}$, the gravitational pull would always be directed toward the opposite foot \cite{Tiseo2018}. The virtual Centers of Rotation (vCoRs) are defined by projecting the segment connecting the CoM and the CoR onto the ground, reflecting recent findings that human motor control relies on a static reference frame within the foot \cite{Carpentier2017, Tiseo2018}.\\
Therefore, the anteroposterior trajectories of the vCoRs (\autoref{fig1}) are described by a cycle that begins when both feet align in the frontal plane. This periodic state occurs every $t_{Step}$ when the CoM reaches its maximum mediolateral oscillation amplitude in the opposite direction (\autoref{eq3}):
\begin{equation}
\label{eq5}
x_{vCoR_i}(t) =
\left\{ \begin{array}{ll}
x_{vCoR_i}(t_0), & \text{from } t_0 \text{ to the end of TO} \\
-d_{SW}/m_{saddle}(t) + x_{vCoR_i}(t_0), & \text{during the swing phase} \\
x_{vCoR_i}(t_0) + d_{SL}, & \text{from HS to the end of the GC}
\end{array}\right. \quad i \in {R, L}
\end{equation}
where $i=\text{R, L}$, $t_0$ is the time at the beginning of the cycle, $m_{saddle}(t)$ is the trajectory of the $y_{Saddle}$ slope (determined from the relative positions of the support foot and the CoM), and GC denotes the gait cycle.\\

\subsection*{Vertical Trajectory of the CoM}
Based on the findings of Carpentier \textit{et al.} \cite{Carpentier2017}, the vertical trajectory of the Centre of Mass (CoM) is modelled as a cycloid curve. To derive the trajectory equation, the vertical oscillation amplitude ($A_z$) and the initial heel-strike time ($t_{\text{HS0}}$) are first determined using Equations~\ref{eq3} and~\ref{eq5}:
\begin{equation}
\label{eq6}
\left\{
\begin{aligned}
& d_{\text{SL}} - d \cos(\theta_{\text{MaxHS}}) - \frac{d_{\text{SW}}\, x_{\text{CoM}}(t_{\text{HS0}})}{y_{\text{CoM}}(t_{\text{HS0}}) - 0.5\, d_{\text{SW}}} = 0 \\[6pt]
& d(\text{CoM}, \text{CoR}_{\text{HS}})_{t_{\text{HS}}} = \sqrt{L_{\text{P0}}^2 - \Delta X_{\text{HS}}^2 - \Delta Y_{\text{HS}}^2} + d \sin(\theta_{\text{MaxHS}} + \theta_{\text{DAH}}) \\[6pt]
& A_z = d(\text{CoM}, \text{CoR}_{\text{Support}})_{t_0} - d(\text{CoM}, \text{CoR}_{\text{HS}})_{t_{\text{HS}}}
\end{aligned}
\right.
\end{equation}
where $d(\cdot, \cdot)_{t}$ denotes the Euclidean distance between two points at time $t$, $\text{CoR}_{\text{HS}}$ is the Centre of Rotation (CoR) of the landing leg, and $\text{CoR}_{\text{Support}}$ is the CoR of the supporting leg passing beneath the CoM. $\Delta X_{\text{HS}}$ and $\Delta Y_{\text{HS}}$ represent the distances between the CoM and $\text{CoR}_{\text{HS}}$ along the anteroposterior and mediolateral axes, respectively. Finally, $t_0$ is the instant at which the swing leg passes the stance leg ($x_{\text{CoR-Support}} = x_{\text{CoR-Swing}}$).

Once these parameters are established, the vertical trajectory of the CoM is given by:
\begin{equation}
\label{eq8}
z_{\text{CoM}}(t) = 
\begin{cases}
(z_{\text{MAX}} - A_z) + A_z \cos\!\left(\pi \dfrac{t + d_{x\text{Ank}}/v_{\text{W}}}{t_{\text{HS0}} + d_{x\text{Ank}}/v_{\text{W}}}\right), & \text{if } t \leq t_{\text{HS0}} \\[10pt]
(z_{\text{MAX}} - A_z) + A_z \cos\!\left(\pi \dfrac{t - t_{\text{HS0}}}{t_{\text{Step}} - t_{\text{HS0}} - d_{x\text{Ank}}/v_{\text{W}}}\right), & \text{if } t_{\text{HS0}} < t \leq t_{\text{Step}} - d_{x\text{Ank}}/v_{\text{W}} \\[10pt]
(z_{\text{MAX}} - A_z) + A_z \cos\!\left(\pi \dfrac{t - t_{\text{Step}} + d_{x\text{Ank}}/v_{\text{W}}}{t_{\text{HS0}} + d_{x\text{Ank}}/v_{\text{W}}}\right), & \text{otherwise}
\end{cases}
\end{equation}
where $z_{\text{MAX}} = \sqrt{L_{\text{P0}}^2 - (0.5\, d_{\text{SW}})^2 - d_{x\text{AH}}^2}$ is the maximum height of the CoM, reached when it aligns vertically with the ankle joint of the supporting foot.

\subsection*{Validation Method}
The model was validated by comparing the mediolateral and vertical CoM amplitudes generated by the proposed framework against the experimental CoM trajectories recorded during the trials. To facilitate this comparison, second-order polynomial regressions were derived for both gait parameters as a function of walking velocity ($v_W$). The regression coefficients were computed at a 95\% confidence level (i.e., mean $\pm$ 2 standard deviations), yielding the following expressions:
\begin{equation}
\label{eq9}
\left\{
\begin{array}{ll}
A_{y-Des} = (0.011 \pm 0.007) v_W^2 - (0.041 \pm 0.022) v_W + (0.051 \pm 0.015) & \text{m} \\
A_{z-Des} = (0.012 \pm 0.013) v_W^2 - (0.003 \pm 0.040) v_W + (0.021 \pm 0.028) & \text{m}
\end{array}\right.
\end{equation}
Furthermore, the duration of the double support phase was analyzed to determine whether an intrinsic regulatory mechanism within the model accounts for the speed-dependent variations in double support duration typically observed in human gait \cite{Orendurff2004, Torricelli2016}.
\section*{Discussion}
Our results demonstrate that the proposed deterministic model, which utilizes a harmonic oscillator centered at the saddle point, successfully produces human-like gait trajectories for both the Center of Mass (CoM) and the swing foot. Furthermore, the model identifies ankle strategies as the primary mechanism for controlling vertical CoM movement, simultaneously regulating walking speed and stability. These findings corroborate and extend prior results obtained from different datasets \cite{Tiseo2016,Tiseo2018, Tiseo2019}.\\
The experimental data indicate that humans regulate step length, step frequency, and mediolateral amplitudes according to a priori optimized behaviors, aligning with observations by Collins \textit{et al.} \cite{Collins2013}. In contrast, parameters such as step width ($d_{SW}$) and vertical CoM amplitude (via ankle strategies) appear to be optimized continuously based on local environmental conditions. This distinction is supported by multiple studies highlighting the modulation of these parameters in response to environmental perturbations \cite{Torricelli2016,Ahn2012,Hak2013}. The critical role of ankle strategies in gait energetic efficiency is also widely documented \cite{Ahn2012,Asbeck2014,Collins2013,Collins2015,Ding2017,Kim2015,MyungheeKim2013,Kuo2007}. Crucially, the proposed model intrinsically captures the regulation of the vertical CoM trajectory and the variations in double support duration observed in human gait \cite{Orendurff2004,wu2019mechanics,Lacquaniti2012}. As noted by Orendurff \textit{et al.}, human trajectories are asymmetric at low speeds due to the anticipation of the minimum CoM height during the gait cycle. By leveraging the Saddle Space proposed by Tiseo \textit{et al.} \cite{Tiseo2016, Tiseo2018,Tiseo2018bioinspired,Tiseo2018b,Tiseo2018deployment}, the current model provides a complete characterization of the bipedal potential energy, which can be derived directly from the CoM trajectory. Consequently, it offers a comprehensive mathematical model of the attractor generated by the interaction between the CoM and gravity.
\subsection*{The Strange Attractor}
\begin{figure}[ht]
\centering
\includegraphics[width=\linewidth]{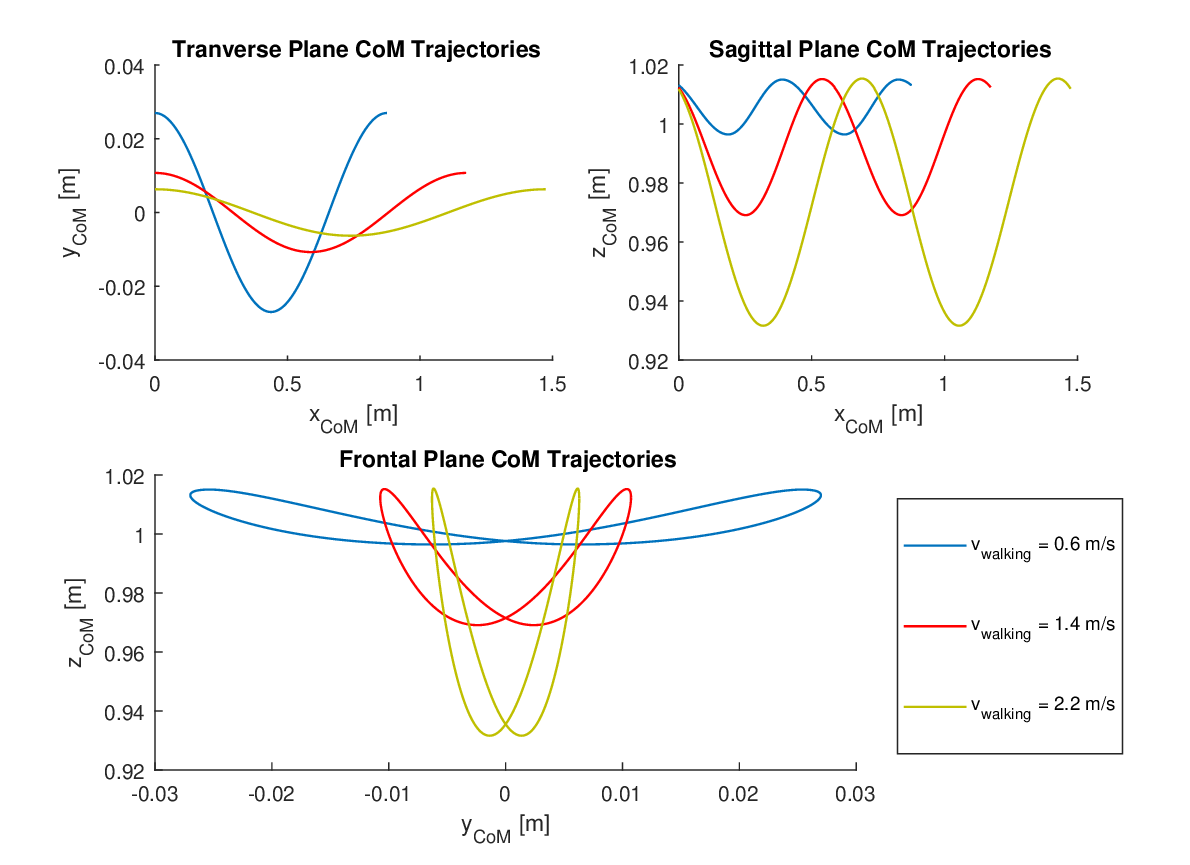}
\caption{CoM trajectories generated by the proposed planner in the transverse, sagittal, and frontal planes for walking speeds ranging from 0.6 to 2.2 m/s. Notably, the frontal view exhibits the typical butterfly shape characteristic of a strange attractor. Furthermore, the proposed model successfully replicates the speed-dependent behavior observed in vivo by Orendurff \textit{et al.}: as walking speed increases, the vertical amplitude of the CoM trajectory increases, the mediolateral amplitude decreases, and the CoM reaches its minimum height closer to zero.}
\label{fig2}
\end{figure}

Human locomotion exhibits significant variability due to neural noise, effector noise, and external perturbations, making it highly improbable for two consecutive steps to be identical. Consequently, while human locomotion appears to be governed by a deterministic planner, the chaotic nature of local conditions dictates that trajectories originating from similar initial states will eventually diverge. It is this chaotic task environment that transforms the deterministic behavior of our proposed model into the strange attractor (centered at the saddle point) observed in empirical human data.\\
This phenomenon is evident when comparing the idealized data from our model (\autoref{fig2}) with the experimental data of a single subject (\autoref{fig8}). Although the human trajectories remain consistent with the behavior predicted by the proposed method, their susceptibility to local conditions introduces an intrinsic variability when tracking the ideal path. While Kang \textit{et al.} \cite{Kang2015} and Orendurff \textit{et al.} \cite{Orendurff2004} previously observed stereotypical dynamics indicative of a strange attractor, they did not explicitly link these observations to a saddle-point attractor model.
\subsubsection*{Implications for Control Architectures}
The intrinsic unpredictability of long-term environmental conditions poses a significant challenge for human motor control, particularly given that transcortical feedback loops require at least 200 ms \cite{Winter1995,Wise2002,Maki2007}. This delay precludes the brain from exerting extensive, continuous control over step-to-step transitions. To explain how the Central Nervous System (CNS) mitigates these physiological limitations, Ahn \textit{et al.} \cite{Ahn2012} proposed a decentralized, semi-autonomous architecture. Our model confirms their framework by identifying ankle strategies as the primary mechanism for regulating local locomotion stability and energy efficiency. Ahn \textit{et al.} further suggested that ankle behaviors are synchronized by a non-linear oscillator, a role fulfilled by the strange attractor identified in our proposed model.\\
We have recently proposed that the CNS relies on two distinct models of the Base of Support (BoS) to plan and supervise locomotion \cite{Tiseo2016, Tiseo2018,Tiseo2018bioinspired}. The Instantaneous-BoS (IBoS) is theorized to function as a region of finite-time invariance used by the cerebellum to supervise Central Pattern Generators (CPGs) in the brainstem and spinal cord. A region of finite-time invariance is defined as a space surrounding the CoM where non-linear, time-dependent dynamics can be considered momentarily invariant \cite{Tiseo2018,Tiseo2018bioinspired}. Common in control systems operating under highly variable conditions, this approach does not guarantee exact convergence to a desired output but provides a robust strategy for maintaining local stability. Furthermore, we hypothesize the existence of a Ballistic-BoS (BBoS), which describes the expected dynamic conditions for a given posture. The BBoS acts as a forward model, implemented by the cerebellum, to predict the future global stability conditions required by the motor cortex to plan stable movements.
\subsection*{An Integrated Motor Control Theory}
Before detailing how our findings synthesize prominent motor control theories, we summarize the primary deductions derived from this work:
\begin{itemize}
\item Step length and lateral CoM amplitude are optimized based on a priori optimal strategies.
\item Step width and vertical CoM amplitude are dynamically adjusted based on local conditions.
\item These findings strongly support a hierarchical architecture composed of semi-autonomous controllers.
\item Ankle strategies are the primary mechanism for controlling vertical CoM movement and appear to be governed by spinal CPGs under cerebellar supervision.
\item The motor cortex, involved in both task-space and joint-space planning, evaluates stability using the IBoS and BBoS models.
\item The task-space planner proposed by our model utilizes the principal direction of a harmonic oscillator centered at the saddle point, rendering it highly compatible with dynamic primitives theory.
\item Desired foot placement emerges from a combination of stereotyped optimal behaviors and dynamic stability requirements dictated by walking speed and environmental conditions.
\end{itemize}
\subsubsection*{Bridging Dynamic and Motor Primitives}
Dynamic motion primitives were initially developed as an extension of motor primitives, yet the exact theoretical connection between them has remained ambiguous \cite{Hogan2012}. Our framework bridges this gap by integrating the proposed task-space planner with the joint-space planner, $\lambda_0$-PMP, discussed in the introduction. In its standard architecture, $\lambda_0$-PMP assumes that task-space dynamics are governed solely by a spring-based mechanism drawing the user toward a desired posture \cite{Tommasino2017}. However, if the forces generated by our proposed saddle attractor replace this virtual spring, the planning process optimizes an energy function that concurrently accounts for desired behavior, mechanical body impedance, and environmental dynamics \cite{Tommasino2017a}.\\
Consequently, the modified $\lambda_0$-PMP architecture produces a sequence of optimized postures fully contextualized within environmental dynamics. CPGs then utilize this sequence of reference postures to generate the neural signals driving muscular activity. This hypothesis is supported by in vivo animal studies, which demonstrate that gait-like behaviors can be induced in rats with spinal injuries via localized electrochemical stimulation of the spinal cord \cite{VandenBrand2012, Minev2015}. Crucially, this integrated architecture implies that when environmental dynamics are negligible, task planning relies exclusively on the reward/discomfort functions of the $\lambda_0$-PMP. Therefore, motor primitives are simply dynamic primitives operating in conditions where intrinsic mechanical body impedance dominates the task.\\
Finally, as illustrated in \autoref{fig9}, this architecture relies on local postural optimization ($\lambda_0$-PMP) acting upon globally optimal strategies provided by the Saddle Task-Space Planner. This implies that the output converges to an optimal strategy provided the body can adequately compensate for environmental variability. As a result, the output of the proposed hierarchical architecture can converge to solutions predicted by Optimal Feedback Control (OFT) theories, but at a significantly lower computational cost.
\subsubsection*{Anatomophysiological Parallels}
A review of motor control literature reveals striking parallels between our proposed control architecture and current understandings of neuroanatomy and physiology. It must be noted, however, that the proposed model is not a literal anatomical map of the CNS, but rather a computational architecture that captures its functional characteristics.\\
The hierarchical, semi-autonomous controller we propose (\autoref{fig9}) mirrors biological motor control pathways. At the apex is the Task-Space (TS) planner, which processes the high-level intention to walk toward a target. TS planning appears to rely on stereotyped, optimized strategies modulated by environmental conditions. Functionally, this structure must: receive voluntary input, assess environmental and somatic conditions, select an optimal strategy from a priori models, and generate desired task-space trajectories and expected external dynamics. Anatomically, these functions align with the frontal and prefrontal cortices, supported by the basal ganglia (regulating reward circuits) and the cerebellum (providing internal planning models via the thalamus and parietal cortex) \cite{Wise2002, Shadmehr2008, Shadmehr2017a, Flash2016}.\\
This information is forwarded to the TS planner, bridging the pre-motor and motor cortices. The motor cortex is subsequently tasked with joint-space planning, aided by the parietal cortex and cerebellum \cite{Wise2002, Shadmehr2008}. The resulting output descends through the brainstem to the spinal cord, where CPGs and spinal reflexes implement the final stages of the CNS architecture \cite{Wise2002, Shadmehr2008}. This spinal stage provides the first centralized response to external perturbations, featuring an ankle reaction time of approximately 80 ms \cite{Maki2007, Ahn2012}. The final tier of this architecture is the intrinsic mechanical impedance of the musculoskeletal system, which naturally filters external perturbations and is modulated by the CNS to execute the planned movements.\\
\subsection*{Implications for Mobility Impairment Rehabilitation}
Our findings demonstrate that human locomotor strategies leverage gravitational forces to maximize energy efficiency, with ankle strategies playing the dominant role in managing both local stability and energy expenditure. This suggests that the low efficacy of many current robotic rehabilitation therapies stems from their high mechanical impedance. By enforcing rigid kinematic trajectories, traditional assistive devices override the natural gravitational dynamics the CNS relies on, effectively suppressing the intrinsic strange attractor required to adapt to local environmental conditions.\\
We propose that the next generation of assistive medical technologies must move away from strict trajectory tracking and instead prioritize dynamically transparent systems driven by advanced interaction control. Rather than fighting the physics of daily living, these devices should modulate mechanical impedance dynamically—providing only the minimal assistance necessary to maintain global postural bounds (conceptually akin to the $\lambda_0$-PMP framework), coupled with highly responsive, minimally invasive robotic actuation specifically targeting ankle strategies. While isolated robotic devices capable of providing such localized support exist \cite{Collins2015, Ding2017, Kim2015, Asbeck2014}, a fully integrated, interaction-driven solution that allows patients to safely exploit natural gravitational attractors has yet to be realized.
\subsection*{Limitations and Future Work}
The primary limitation of this work is its exclusive focus on straight-line locomotion over flat, unobstructed surfaces. While the dataset of unperturbed walking effectively demonstrates the model's baseline validity, it does not explicitly test the system's response to environmental unpredictability. For instance, while the remarkably low correlation ($R^2$) between walking velocity and parameters such as step width or heel-strike angle strongly suggests these variables are modulated by local stability demands, our current dataset lacks the physical perturbations (e.g., uneven terrain, lateral pushes, or varying ground impedances) required to definitively prove this causal link.\\
Furthermore, locomotion can be viewed as the antipode of reaching tasks: it is dominated by gravitational attractor dynamics rather than intrinsic mechanical body impedance. Consequently, several critical questions must be addressed before this unified theory can be fully validated:
\begin{itemize}
\item How does the CNS balance desired intentional behavior with dominant environmental dynamics? (i.e., what control policies govern the middle ground between pure locomotion and upper-limb reaching tasks?)
\item How does the CNS regulate the transition between control policies to adapt global strategies to rapidly changing environmental conditions?
\item How does the proposed hierarchical architecture perform under explicit physical perturbations, and can rigorous quantitative metrics (such as peak positional error or RMSE) during these interactions empirically validate the hypothesized role of ankle strategies?
\end{itemize}
Lastly, while the anatomophysiological parallels draw a compelling functional bridge between our kinematic limit-cycle observations and internal CNS structures (such as cerebellar forward models), the leap from behavioral kinematic data to internal neural architecture remains theoretical. As with all computational motor control theories, this framework does not prove that the human brain explicitly computes these specific mathematical formulations. Rather, it suggests that the proposed decentralized controller serves as a highly accurate, computationally efficient analog for describing human motor control, offering a promising blueprint for biomimetic robotic architectures.

\section*{Acknowledgements}
This work is partially extracted from the PhD Thesis of Carlo Tiseo \cite{Tiseo2018}. This research was supported by the A*STAR-NHG-NTU Rehabilitation Research Grant: "Mobile Robotic Assistive Balance Trainer" (RRG/16018).  

\section*{Additional information}

None of the authors has competing financial interests.

\bibliography{library}

\begin{thebibliography}{10}

\bibitem{inman1953major}
Verne~T Inman, Howard~D Eberhart, et~al.
\newblock The major determinants in normal and pathological gait.
\newblock {\em Jbjs}, 35(3):543--558, 1953.

\bibitem{duysens2019controller}
Jacques Duysens and Arturo Forner-Cordero.
\newblock A controller perspective on biological gait control: Reflexes and central pattern generators.
\newblock {\em Annual Reviews in Control}, 48:392--400, 2019.

\bibitem{duysens2018walking}
Jacques Duysens and Arturo Forner-Cordero.
\newblock Walking with perturbations: a guide for biped humans and robots.
\newblock {\em Bioinspiration \& Biomimetics}, 13(6):061001, 2018.

\bibitem{Torricelli2016}
Diego Torricelli, Jose Gonzalez, Maarten Weckx, Ren{\'{e}} Jim{\'{e}}nez-Fabi{\'{a}}n, Bram Vanderborght, Massimo Sartori, Strahinja Dosen, Dario Farina, Dirk Lefeber, and Jose~L Pons.
\newblock {Human-like compliant locomotion: state of the art of robotic implementations}.
\newblock {\em Bioinspiration {\&} Biomimetics}, 11(5):051002, aug 2016.

\bibitem{Hogan2012}
Neville Hogan and Dagmar Sternad.
\newblock {Dynamic primitives of motor behavior}.
\newblock {\em Biological Cybernetics}, 106(11-12):727--739, dec 2012.

\bibitem{Hogan2013}
Neville Hogan and Dagmar Sternad.
\newblock {Dynamic primitives in the control of locomotion}.
\newblock {\em Frontiers in Computational Neuroscience}, 7, 2013.

\bibitem{Winter1995}
DA~Winter.
\newblock {Human balance and posture control during standing and walking}.
\newblock {\em Gait {\&} Posture}, 3(4):193--214, dec 1995.

\bibitem{Ivanenko2006}
Yuri~P Ivanenko, Richard~E. Poppele, and Francesco Lacquaniti.
\newblock {Motor Control Programs and Walking}.
\newblock {\em The Neuroscientist}, 12(4):339--348, aug 2006.

\bibitem{Kuo2007}
Arthur~D Kuo.
\newblock {The six determinants of gait and the inverted pendulum analogy: A dynamic walking perspective}.
\newblock {\em Human Movement Science}, 26(4):617--656, aug 2007.

\bibitem{Tommasino2017}
Paolo Tommasino and Domenico Campolo.
\newblock {An Extended Passive Motion Paradigm for Human-Like Posture and Movement Planning in Redundant Manipulators}.
\newblock {\em Frontiers in Neurorobotics}, 11(November):1--17, 2017.

\bibitem{Lacquaniti2012}
Francesco Lacquaniti, Yuri~P Ivanenko, and Myrka Zago.
\newblock {Patterned control of human locomotion}.
\newblock {\em The Journal of Physiology}, 590(10):2189--2199, may 2012.

\bibitem{Zelik2014}
Karl~E Zelik, Valentina {La Scaleia}, Yuri~P Ivanenko, and Francesco Lacquaniti.
\newblock {Can modular strategies simplify neural control of multidirectional human locomotion?}
\newblock {\em Journal of Neurophysiology}, 111(8):1686--1702, apr 2014.

\bibitem{Flash2016}
Tamar Flash and Emilio Bizzi.
\newblock {Cortical circuits and modules in movement generation: experiments and theories}.
\newblock {\em Current Opinion in Neurobiology}, 41:174--178, dec 2016.

\bibitem{Ahn2012}
Jooeun Ahn and Neville Hogan.
\newblock {Walking Is Not Like Reaching: Evidence from Periodic Mechanical Perturbations}.
\newblock {\em PLoS ONE}, 7(3):e31767, mar 2012.

\bibitem{Tiseo2018}
Carlo Tiseo.
\newblock {\em {Modelling of bipedal locomotion for the development of a compliant pelvic interface between human and a balance assistant robot}}.
\newblock PhD thesis, Nanyang Technological University, 2018.

\bibitem{Shadmehr2008}
Reza Shadmehr and John~W Krakauer.
\newblock {A computational neuroanatomy for motor control}.
\newblock {\em Experimental Brain Research}, 185(3):359--381, mar 2008.

\bibitem{Ajemian2010}
Robert Ajemian and Neville Hogan.
\newblock {Experimenting with Theoretical Motor Neuroscience}.
\newblock {\em Journal of Motor Behavior}, 42(6):333--342, oct 2010.

\bibitem{hausdorff2005gait}
Jeffrey~M Hausdorff.
\newblock Gait variability: methods, modeling and meaning.
\newblock {\em Journal of neuroengineering and rehabilitation}, 2(1):1--9, 2005.

\bibitem{delignieres2009fractal}
Didier Deligni{\`e}res and Kjerstin Torre.
\newblock Fractal dynamics of human gait: a reassessment of the 1996 data of hausdorff et al.
\newblock {\em Journal of Applied Physiology}, 106(4):1272--1279, 2009.

\bibitem{forner2006describing}
A~Forner-Cordero, HJFM Koopman, and FCT Van Der~Helm.
\newblock Describing gait as a sequence of states.
\newblock {\em Journal of biomechanics}, 39(5):948--957, 2006.

\bibitem{Tiseo2016}
Carlo Tiseo and Wei~Tech Ang.
\newblock {The Balance: An energy management task}.
\newblock In {\em 2016 6th IEEE International Conference on Biomedical Robotics and Biomechatronics (BioRob)}, pages 723--728. IEEE, jun 2016.

\bibitem{tiseo2018bipedal}
C~Tiseo, KC~Veluvolu, and WT~Ang.
\newblock The bipedal saddle space: modelling and validation.
\newblock {\em Bioinspiration \& biomimetics}, 14(1):015001, 2018.

\bibitem{Tiseo2018bioinspired}
Carlo Tiseo, Kalyana~C Veluvolu, and Wei~Tech Ang.
\newblock Motor control insights on walking planner and its stability.
\newblock {\em Engineering Research Express}, 5(2):025009, 2023.

\bibitem{Tiseo2018a}
Carlo Tiseo, Ming~Jeat Foo, Kalyana~C Veluvolu, and Wei~Tech Ang.
\newblock {A Postural Model for Tracking the Base of Support}.
\newblock In {\em 2018 40th Annual International Conference of the IEEE Engineering in Medicine and Biology Society, EMBC 2018}, 2018.

\bibitem{Mandery2015}
Christian Mandery, Omer Terlemez, Martin Do, Nikolaus Vahrenkamp, and Tamim Asfour.
\newblock {The KIT whole-body human motion database}.
\newblock In {\em 2015 International Conference on Advanced Robotics (ICAR)}, pages 329--336. IEEE, jul 2015.

\bibitem{Hak2013}
Laura Hak, Han Houdijk, Peter~J. Beek, and Jaap~H. van Die{\"{e}}n.
\newblock {Steps to Take to Enhance Gait Stability: The Effect of Stride Frequency, Stride Length, and Walking Speed on Local Dynamic Stability and Margins of Stability}.
\newblock {\em PLoS ONE}, 8(12):e82842, dec 2013.

\bibitem{Jo2007}
Sungho Jo and Steve~G Massaquoi.
\newblock {A model of cerebrocerebello-spinomuscular interaction in the sagittal control of human walking}.
\newblock {\em Biological Cybernetics}, 96(3):279--307, mar 2007.

\bibitem{Virmavirta2014}
Mikko Virmavirta and Juha Isolehto.
\newblock {Determining the location of the body's center of mass for different groups of physically active people}.
\newblock {\em Journal of Biomechanics}, 47(8):1909--1913, jun 2014.

\bibitem{Tiseo2018b}
Carlo Tiseo, Kalyana~C Veluvolu, and Wei~Tech Ang.
\newblock {Evidence of a “ Clock ” Determining Human Locomotion}.
\newblock In {\em 2018 40th Annual International Conference of the IEEE Engineering in Medicine and Biology Society, EMBC 2018}, 2018.

\bibitem{Carpentier2017}
Justin Carpentier, Mehdi Benallegue, and Jean-Paul Laumond.
\newblock {On the centre of mass motion in human walking}.
\newblock {\em International Journal of Automation and Computing}, 2017.

\bibitem{Orendurff2004}
Michael~S Orendurff, Ava~D Segal, Glenn~K Klute, Jocelyn~S Berge, Eric~S Rohr, and Nancy~J Kadel.
\newblock {The effect of walking speed on center of mass displacement.}
\newblock {\em Journal of rehabilitation research and development}, 41(6A):829--34, 2004.

\bibitem{Tiseo2019}
Carlo Tiseo, Sethu Vijayakumar, and Michael Mistry.
\newblock Analytic model for quadruped locomotion task-space planning.
\newblock In {\em 2019 41th Annual International Conference of the IEEE Engineering in Medicine and Biology Society, EMBC 2019}, 2019.

\bibitem{Collins2013}
Steven~H. Collins and Arthur~D. Kuo.
\newblock {Two Independent Contributions to Step Variability during Over-Ground Human Walking}.
\newblock {\em PLoS ONE}, 8(8):e73597, aug 2013.

\bibitem{Asbeck2014}
Alan~T Asbeck, Stefano~M.M. {De Rossi}, Ignacio Galiana, Ye~Ding, and Conor~J Walsh.
\newblock {Stronger, Smarter, Softer: Next-Generation Wearable Robots}.
\newblock {\em IEEE Robotics {\&} Automation Magazine}, 21(4):22--33, dec 2014.

\bibitem{Collins2015}
Steven~H. Collins, M.~Bruce Wiggin, and Gregory~S. Sawicki.
\newblock {Reducing the energy cost of human walking using an unpowered exoskeleton}.
\newblock {\em Nature}, 522(7555):212--215, apr 2015.

\bibitem{Ding2017}
Ye~Ding, Ignacio Galiana, Alan~T. Asbeck, Stefano Marco~Maria {De Rossi}, Jaehyun Bae, Thiago Ribeiro~Teles Santos, Vanessa~Lara {De Araujo}, Sangjun Lee, Kenneth~G. Holt, and Conor Walsh.
\newblock {Biomechanical and physiological evaluation of multi-joint assistance with soft exosuits}.
\newblock {\em IEEE Transactions on Neural Systems and Rehabilitation Engineering}, 25(2):119--130, 2017.

\bibitem{Kim2015}
Myunghee Kim and Steven~H Collins.
\newblock {Once-per-step control of ankle-foot prosthesis push-off work reduces effort associated with balance during walking}.
\newblock {\em Journal of NeuroEngineering and Rehabilitation}, 12(1):43, dec 2015.

\bibitem{MyungheeKim2013}
{Myunghee Kim} and Steven~H. Collins.
\newblock {Stabilization of a three-dimensional limit cycle walking model through step-to-step ankle control}.
\newblock In {\em 2013 IEEE 13th International Conference on Rehabilitation Robotics (ICORR)}, pages 1--6. IEEE, jun 2013.

\bibitem{wu2019mechanics}
Amy~R Wu, Cole~S Simpson, Edwin~HF van Asseldonk, Herman van~der Kooij, and Auke~J Ijspeert.
\newblock Mechanics of very slow human walking.
\newblock {\em Scientific reports}, 9(1):18079, 2019.

\bibitem{Tiseo2018deployment}
Carlo Tiseo, Ming~Jeat Foo, Kalyana~C Veluvolu, and Wei~Tech Ang.
\newblock Deployment of the saddle space transformation in tracking the base of support.
\newblock {\em arXiv preprint arXiv:1805.09456}, 2018.

\bibitem{Kang2015}
Jiyeon Kang, Vineet Vashista, and Sunil~K Agrawal.
\newblock {A novel assist-as-needed control method to guide pelvic trajectory for gait rehabilitation}.
\newblock In {\em 2015 IEEE International Conference on Rehabilitation Robotics (ICORR)}, pages 630--635. IEEE, aug 2015.

\bibitem{Wise2002}
Steven~P. Wise and Reza Shadmehr.
\newblock {Motor Control}.
\newblock In {\em Encyclopedia of the Human Brain}, volume~3, pages 137--157. Elsevier, 2002.

\bibitem{Maki2007}
B~E Maki and W~E McIlroy.
\newblock {Cognitive demands and cortical control of human balance-recovery reactions}.
\newblock {\em Journal of Neural Transmission}, 114(10):1279--1296, oct 2007.

\bibitem{Tommasino2017a}
Paolo Tommasino and Domenico Campolo.
\newblock {Task-space separation principle: a force-field approach to motion planning for redundant manipulators}.
\newblock {\em Bioinspiration {\&} Biomimetics}, 12(2):026003, feb 2017.

\bibitem{VandenBrand2012}
R~van~den Brand, J.~Heutschi, Q.~Barraud, J.~DiGiovanna, K.~Bartholdi, M.~Huerlimann, L.~Friedli, I.~Vollenweider, E.~M. Moraud, S.~Duis, N.~Dominici, S.~Micera, P.~Musienko, and G.~Courtine.
\newblock {Restoring Voluntary Control of Locomotion after Paralyzing Spinal Cord Injury}.
\newblock {\em Science}, 336(6085):1182--1185, jun 2012.

\bibitem{Minev2015}
Ivan~R Minev, Pavel Musienko, Arthur Hirsch, Quentin Barraud, Nikolaus Wenger, Eduardo~Martin Moraud, J{\'{e}}r{\^{o}}me Gandar, Marco Capogrosso, Tomislav Milekovic, L{\'{e}}onie Asboth, Rafael~Fajardo Torres, Nicolas Vachicouras, Qihan Liu, Natalia Pavlova, Simone Duis, Alexandre Larmagnac, J.~Voros, Silvestro Micera, Zhigang Suo, Gr{\'{e}}goire Courtine, and St{\'{e}}phanie~P Lacour.
\newblock {Electronic dura mater for long-term multimodal neural interfaces}.
\newblock {\em Science}, 347(6218):159--163, jan 2015.

\bibitem{Shadmehr2017a}
Reza Shadmehr.
\newblock {Distinct neural circuits for control of movement vs. holding still}.
\newblock {\em Journal of Neurophysiology}, page jn.00840.2016, 2017.

\end{thebibliography}
\bibliographystyle{unsrt}

\begin{figure}[ht]
	\centering
	\includegraphics[width=\linewidth]{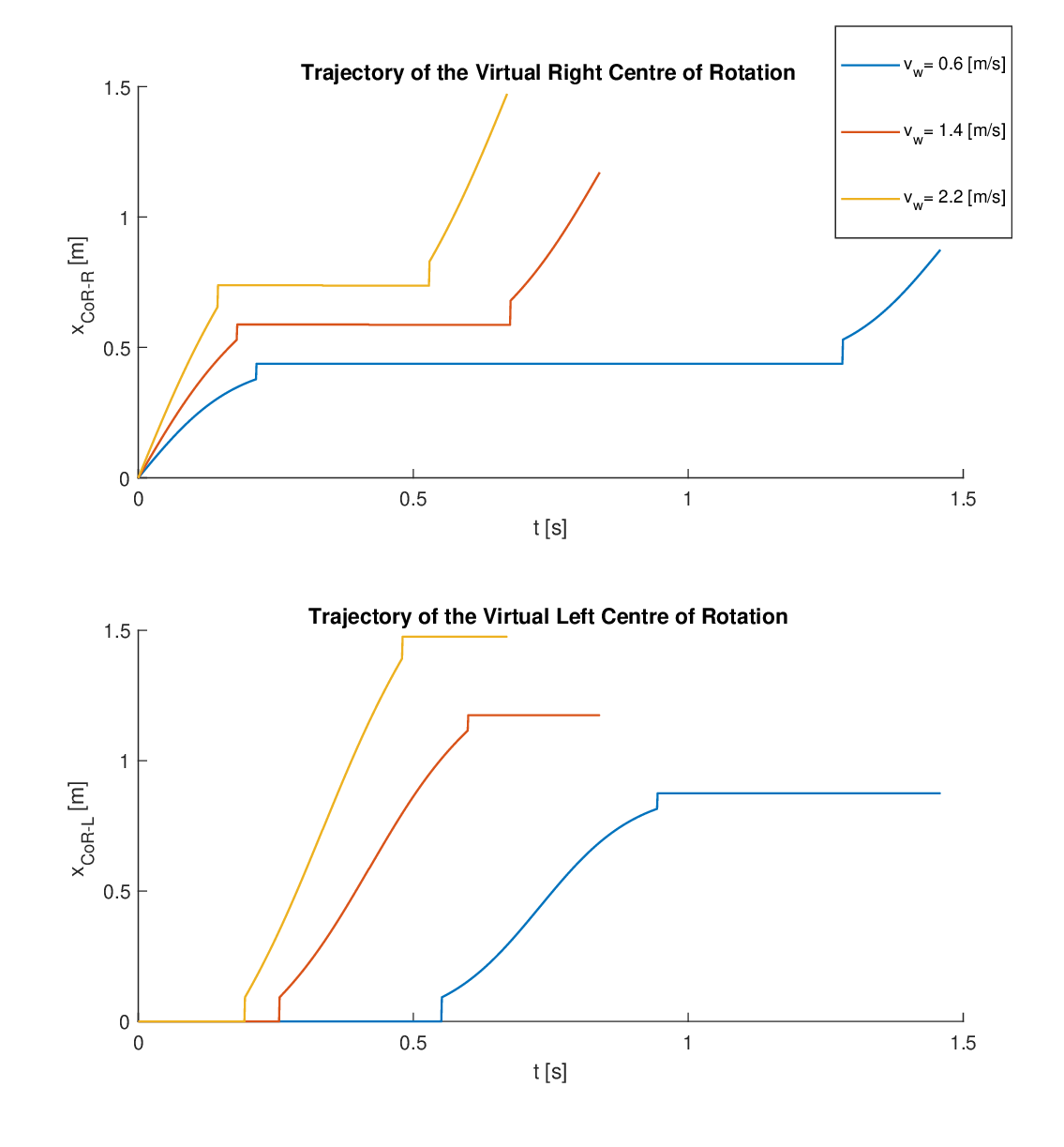}
	\caption{Anteroposterior trajectories of the vCoRs across different walking speeds, demonstrating the intrinsic regulation of the support and swing phases. At lower speeds, the support phase dominates the gait cycle, whereas this dominance diminishes at higher speeds.}
	\label{fig3}
\end{figure}
\begin{figure}[ht]
	\centering
	\includegraphics[width=\linewidth]{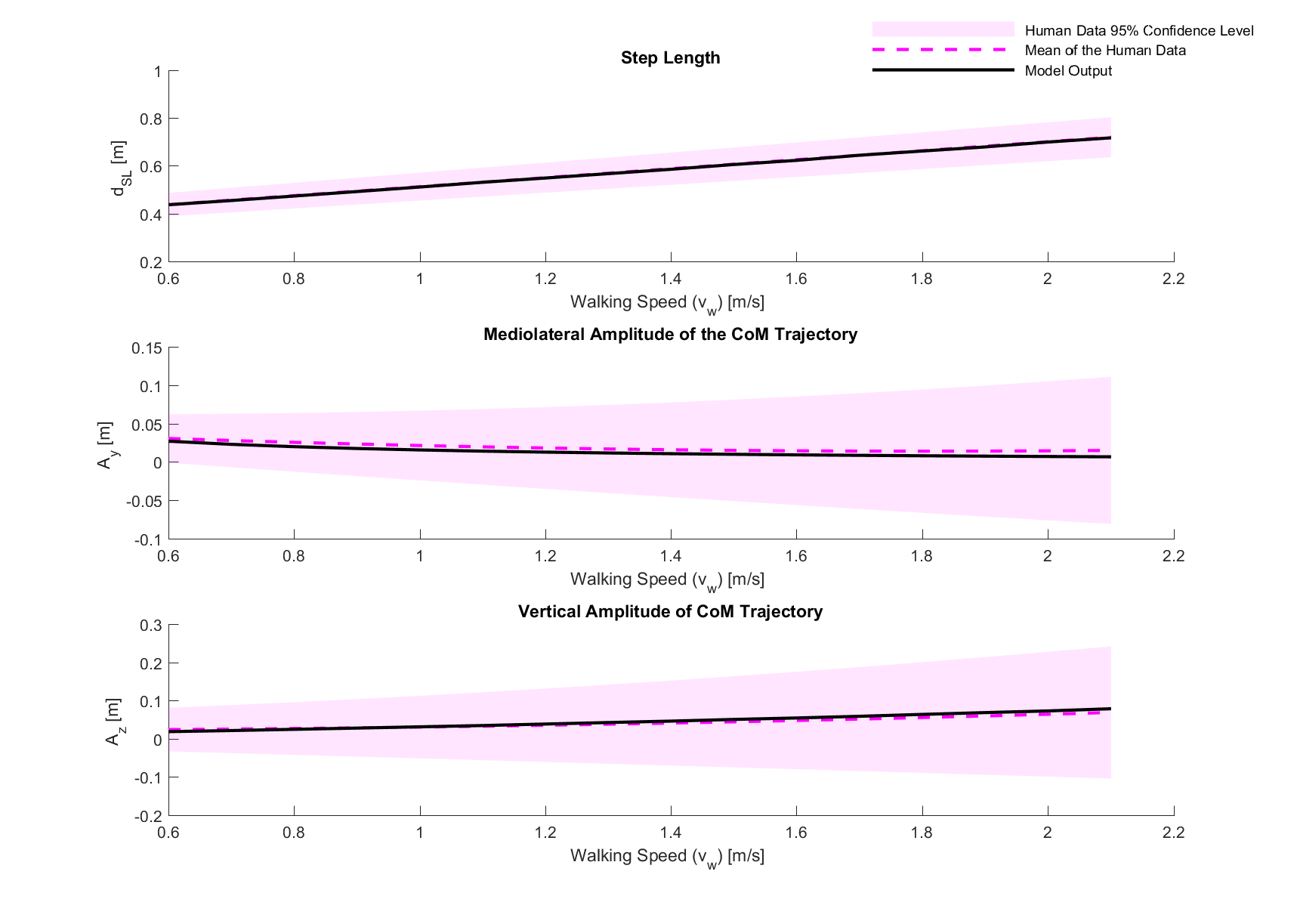}
	\caption{The proposed model generates swing foot, mediolateral, and vertical trajectories that fall within the 95\% confidence intervals of the experimental data, closely matching the mean human behavior.}
	\label{fig4}
\end{figure}
\begin{figure}[ht]
	\centering
	\includegraphics[width=\linewidth,trim=2.5cm 9cm 2cm 9cm,clip]{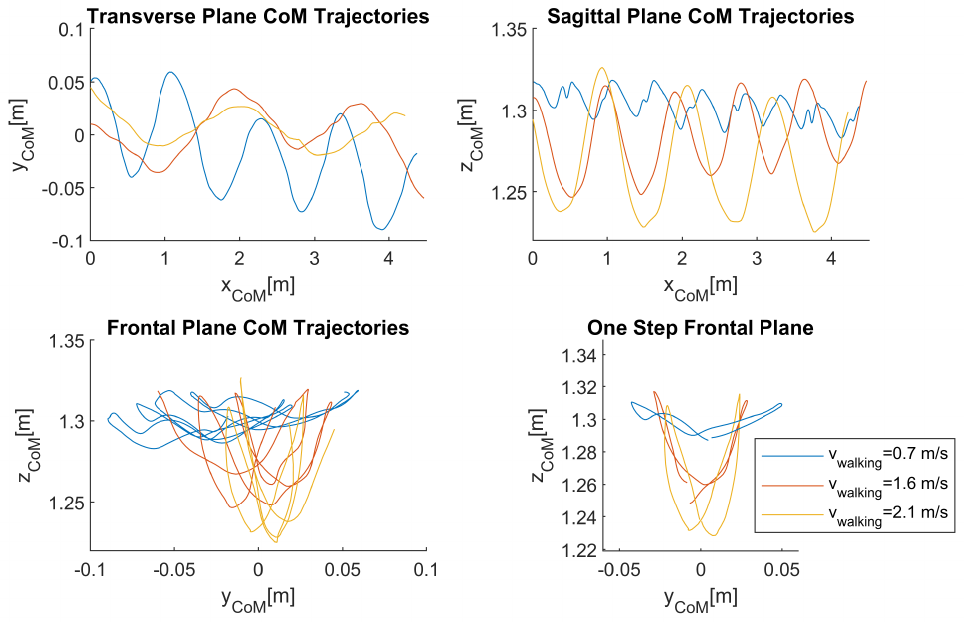}
	\caption{Kinematic trajectories of a single representative subject across the three anatomical planes at slow, normal, and fast walking speeds. These experimental data confirm that the proposed model captures idealized locomotion strategies, which are subsequently modulated by local conditions (e.g., somatic state and environmental interactions). The isolated single-step view in the frontal plane further highlights this phenomenon: speed-dependent trajectory adaptations in the in vivo data closely mirror the simulated model behaviors depicted in \autoref{fig2}.}
	\label{fig8}
\end{figure}
\begin{figure}[ht]
	\centering
	\includegraphics[width=\linewidth]{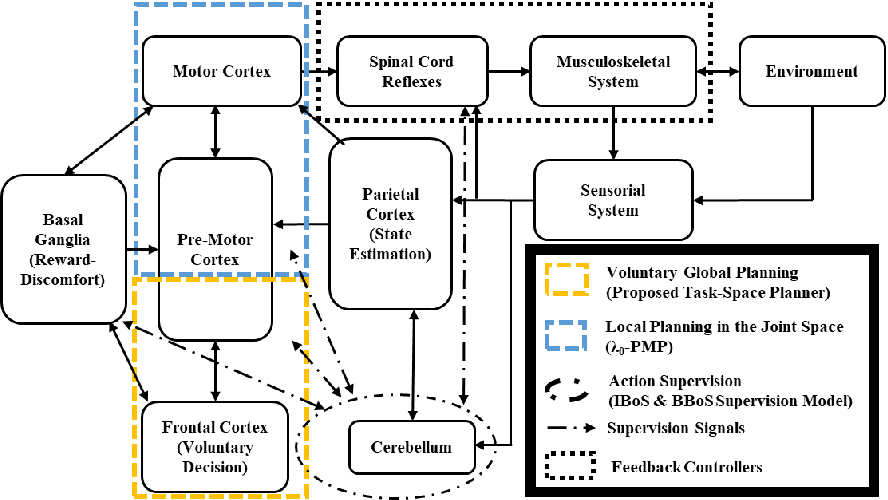}
	\caption{Anatomophysiological parallels between the proposed control architecture and the human central nervous system, synthesized from existing literature \cite{Ajemian2010, Wise2002, Ivanenko2006, Shadmehr2017a, Shadmehr2008, Lacquaniti2012, Flash2016, Hogan2012, Tommasino2017}. Solid lines represent active control information, whereas dotted lines denote supervisory signals. Both types of neural pathways are frequently mediated by additional brain regions not depicted in this simplified schematic. Notably, the pre-motor cortex is positioned across the boundary of the task-space and joint-space planners to reflect our hypothesis that it functions as a functional bridge between the frontal and motor cortices.}
	\label{fig9}
\end{figure}

\begin{table}[ht]
	\centering
	\begin{tabular}{|c||c|c|}
		\hline
		$v_W$ (m/s) & Double Support & Foot Ground Contact \\
		 &  [\% Gait Cycle] & [\% Gait Cycle] \\
		\hline
		\hline
		0.6 & 46 & 73 \\
		\hline
		0.8 & 32 & 66 \\
		\hline
		1 & 24 & 62 \\
		\hline
		1.2 & 20 & 60 \\
		\hline
		1.4 & 18 & 59 \\
		\hline
		1.6 & 16 & 58 \\
		\hline
		1.8 & 16 & 58 \\
		\hline
		2 & 16 & 58 \\
		\hline
		2.2 & 14 & 57 \\
		\hline

	\end{tabular}
	\caption{\label{tab1} The intrinsic regulation of the durations of foot-ground contact and the double support phase emerges directly from the coordinated foot swing and ankle strategies.}
\end{table}

\end{document}